%% file: main.tex
\documentclass[10pt,twocolumn,letterpaper]{article}

\usepackage[pagenumbers]{cvpr}      % To produce the REVIEW version
\input{preamble}

\usepackage{algorithm}
\usepackage{algorithmic}
\usepackage{multirow}
\usepackage{amssymb}
\definecolor{cvprblue}{rgb}{0.21,0.49,0.74}
\usepackage[pagebackref,breaklinks,colorlinks,citecolor=cvprblue]{hyperref}

\def\paperID{15846} % *** Enter the Paper ID here
\def\confName{CVPR}
\def\confYear{2024}
\usepackage{xspace}
\makeatletter
\DeclareRobustCommand\onedot{\futurelet\@let@token\@onedot}
\def\@onedot{\ifx\@let@token.\else.\null\fi\xspace}
\def\eg{\emph{e.g}\onedot} 
\def\ie{\emph{i.e}\onedot}

\usepackage{color}
\definecolor{dark_red}{rgb}{0.5, 0, 0}
\definecolor{dark_green}{rgb}{0, 0.5, 0}
\definecolor{red}{rgb}{.9,0,0}
\definecolor{balck}{rgb}{0,0,0}

\title{Enhancing Zero-shot Counting via Language-guided Exemplar Learning}

\author{Mingjie Wang\\
Zhejiang Sci-Tech University\\
\and
Jun Zhou\\
Dalian Maritime University\\
\and
Yong Dai\\
Tencent AI Lab\\
\and
Eric Buys\\
University of Guelph, Canada\\
\and
Minglun Gong\\
University of Guelph, Canada\\
}

\begin{document}
\maketitle
\input{0_abstract}    
\input{1_intro}

\input{2_review}
\input{3_method}

\input{4_experiments}

{
    \small
    \bibliographystyle{ieeenat_fullname}
    \bibliography{main}
}


\end{document}

%% file: preamble.tex
\usepackage[dvipsnames]{xcolor}

%% file: 0_abstract.tex
\begin{abstract}
Recently, Class-Agnostic Counting (CAC) problem has garnered increasing attention owing to its intriguing generality and superior efficiency compared to Category-Specific Counting (CSC). This paper proposes a novel ExpressCount to enhance zero-shot object counting by delving deeply into language-guided exemplar learning. Specifically, the ExpressCount is comprised of an innovative Language-oriented Exemplar Perceptron and a downstream visual Zero-shot Counting pipeline. Thereinto, the perceptron hammers at exploiting accurate exemplar cues from collaborative language-vision signals by inheriting rich semantic priors from the prevailing pre-trained Large Language Models (LLMs), whereas the counting pipeline excels in mining fine-grained features through dual-branch and cross-attention schemes, contributing to the high-quality similarity learning. Apart from building a bridge between the LLM in vogue and the visual counting tasks, expression-guided exemplar estimation significantly advances zero-shot learning capabilities for counting instances with arbitrary classes. Moreover, devising a FSC-147-Express with annotations of meticulous linguistic expressions pioneers a new venue for developing and validating language-based counting models. Extensive experiments demonstrate the state-of-the-art performance of our ExpressCount, even showcasing the accuracy on par with partial CSC models. 
%The implementation codes will be made publicly available.
\end{abstract}

%It can seamlessly count objects in different settings, be it cells in a biology slide, people in crowd images, or stars in a galaxy, without the need for extensive retraining. 

%% file: 1_intro.tex
\section{Introduction}
\begin{figure}[h]
	\begin{center}
		\includegraphics[width=\linewidth]{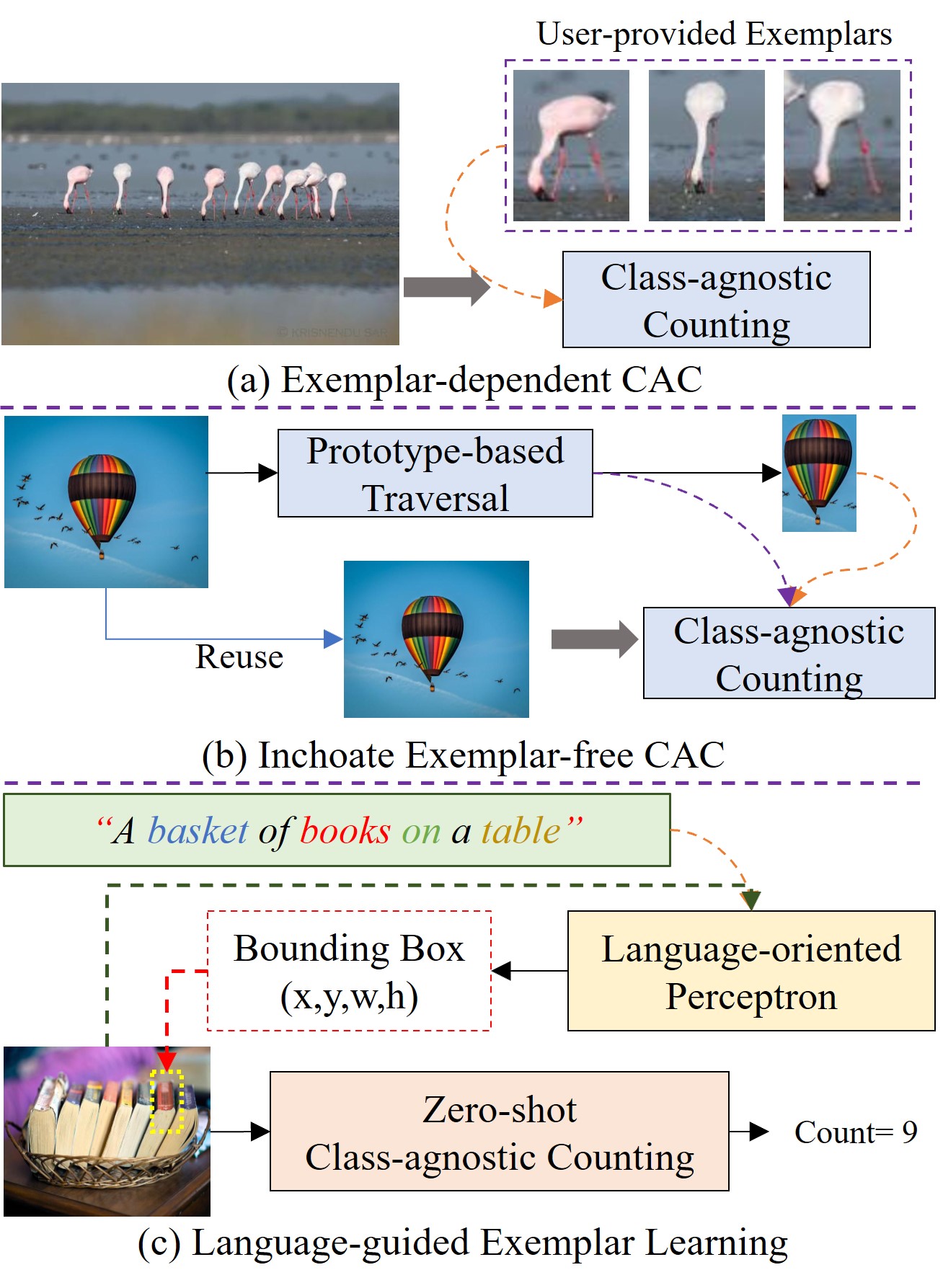}
	\end{center} 
	\caption{(a) The conventional CAC models necessitate user-provided exemplars, imposing a substantial manual burden for specifying dense object locations and impeding the models' applicability; (b) Rudimentary exemplar-free CAC hammers at learning exemplar cues in a traversal manner, often resulting in unsatisfactory results owing to the presence of semantic ambiguity; (c) Language-guided Exemplar Learning excels in enriching linguistic semantics to steer the accurate exemplar/counts regressions. }
	\label{fig:comp1}
\end{figure}

Over the past decade, Class-Specific Counting (CSC) problem has garnered lots of attention from the research community due to its applicability across a wide range of scenarios, such as crowd ~\cite{han2023steerer,liu2023point,huang2023counting,liang2023crowdclip,li2023calibrating,ranasinghe2023diffuse,song2021rethinking,li2018csrnet}, cells~\cite{khan2016deep}, fruits~\cite{rahnemoonfar2017deep} and vehicles~\cite{bui2020vehicle} counting models. Albeit presenting fascinating performance, they fully hinge on the acquisition of massive annotations specific to distinct object classes~(\eg  1.51 million dots for the JHU-CROWD++~\cite{sindagi2020jhu}). 
To reduce the burdensome expenses associated with labeling and widen the versatility of counting methods, a growing number of Class-Agnostic Counting (CAC) models~\cite{chattopadhyay2017counting,lu2018class,yang2021class,ranjan2021learning,shi2022represent,ranjan2022exemplar,lin2022scale,you2023few,jiang2023clip, wang2023gcnet,xu2023zero}  are proposed to count instances distributed throughout various input scenes, regardless of their specific categories. It can seamlessly count objects in different settings, be it cells in a biology slide, people in crowd images, or stars in a galaxy, without the need for extensive retraining. In most cases, these approaches explicitly require user-provided exemplars; see Figure~\ref{fig:comp1}(a), which is impractical for certain real-world applications, like underwater biological analyzing system~\cite{fabic2013fish}, where objects in question are amorphous. 

More recently, researches are transitioning towards a more flexible and intelligent protocol of exemplar-free zero-shot counting, namely Exemplar-Free Counting (EFC)~\cite{ranjan2022exemplar,jiang2023clip, wang2023gcnet,xu2023zero}. However, existing EFC approaches either capture exemplar cues by analyzing the raw image without rich guidance of semantics~\cite{ranjan2022exemplar,wang2023gcnet}; see Figure~\ref{fig:comp1}(b), or recently resort to simple class name for learning discriminative object features~\cite{xu2023zero,jiang2023clip}. 
Although previous efforts have been made to pioneer to expel the nuisance of demanding exemplars annotated by users, the following drawbacks impede their applicability: 
\emph{i)} All current EFC models simply take as input raw image/one-word name to guide the exemplar mining, neglecting the impetus of rich semantics from natural language~\cite{shen2023text,wang2023imagen}; \emph{ii)} The surging of Large Language Models (LLMs)~\cite{devlin2018bert,brown2020language,chang2023survey,zhao2023survey} has paved the way for advancing performance of various vision tasks~\cite{zhao2023learning,koh2023grounding,zhong2023adapter}, thanks to powerful transferability of LLMs. Nevertheless, CAC has yet tap into advantages offered by LLMs for enhancing zero-shot learning capability; \emph{iii)} Inchoate EFC tends to extract exemplar representations by scanning the entire input image, a process that is both resource-intensive and less adept at quickly identifying location hints. Conversely, as noted by Dehaene~\cite{dehaene2011number}, humans excel at quickly locating the objects by comprehending linguistic signals (\eg language and texts) before embarking on downstream counting tasks.

%, see Figure~\ref{fig:comp2}.

%\begin{figure}[t]
%	\begin{center}
%		\includegraphics[width=\linewidth]{comp2.jpg}
%	\end{center} 
%	\caption{Our proposed linguistics-centered protocol where the counter is capable of parsing human language and predicting accurate exemplars. \mlc{I felt the figure is a bit misleading.  Does your algorithm really take textual questions and determine the counting objects accordingly?  Also, the arrow toward BLIP seems to be a generic prompt for describing the scene, rather than a counting-specific prompting.} } 
%	\label{fig:comp2}
%\end{figure}

To remedy these deficiencies, a novel expression-guided CAC model, namely \emph{ExpressCount}, is proposed to enhance the performance of zero-shot counting by enriching semantic priors from natural expressions; see Figure~\ref{fig:comp1}(c). We aim to endow the EFC with human-like language comprehension. For instance, when presented with the expression ``a basket of books on a table'' along with an input image, humans, leveraging their accumulated knowledge priors, can effortlessly discern rich semantic information, such as the objects distribution dispersing across a 2D scene of ``books in a basket'' and the rough quantity perception of ``books''.
Specifically, a language-oriented exemplar perceptron is presented to predict precise exemplars by mining rich linguistic semantic priors. To the best of our knowledge, this marks the inaugural endeavour to construct a bridge between the LLMs and visual counting problems. However, the foremost challenge stems from the absence of annotated language-vision pairs, thereby hindering the smooth training of language-compatible modules. To surmount this obstacle, we extend the largest CAC-specific FSC-147~\cite{ranjan2021learning} in vogue, by outfitting each sample with fine-grained expressions. The descriptions are crafted and fine-tuned using the pre-trained visual-language BLIP~\cite{li2022blip} and GPT~\cite{openai2023gpt4}. This enriched dataset, named \emph{FSC-147-Express}, provides a new venue for studying language-oriented counting methods.
%\mj{Building upon the language-oriented exemplar perceptron and the extended FSC-147-Express dataset, our ExpressCount allows for inferring exemplar cues (such as location and scale of target objects) from \ml{input scenes and the associated language descriptions}. Then, the downstream zero-shot counting module takes as input the learned exemplar patch and raw image to accomplish the instance counting with target class. By leveraging the guidance of semantic priors and zero-shot ability from pre-trained language and visual models, we achieve more accurate exemplar prediction, as opposed to the undirected learning in existing approaches. 
Extensive experiments demonstrate the state-of-the-art performance of our proposed ExpressCount, emphasizing the significance of annotating language for counting problems. In a nutshell, this work offers four contributions:
%It not only surpasses current EFC approaches in performance, but also competes effectively with exemplar-dependent counters.

\begin{figure*}[h]
	\begin{center}
		\includegraphics[width=\linewidth]{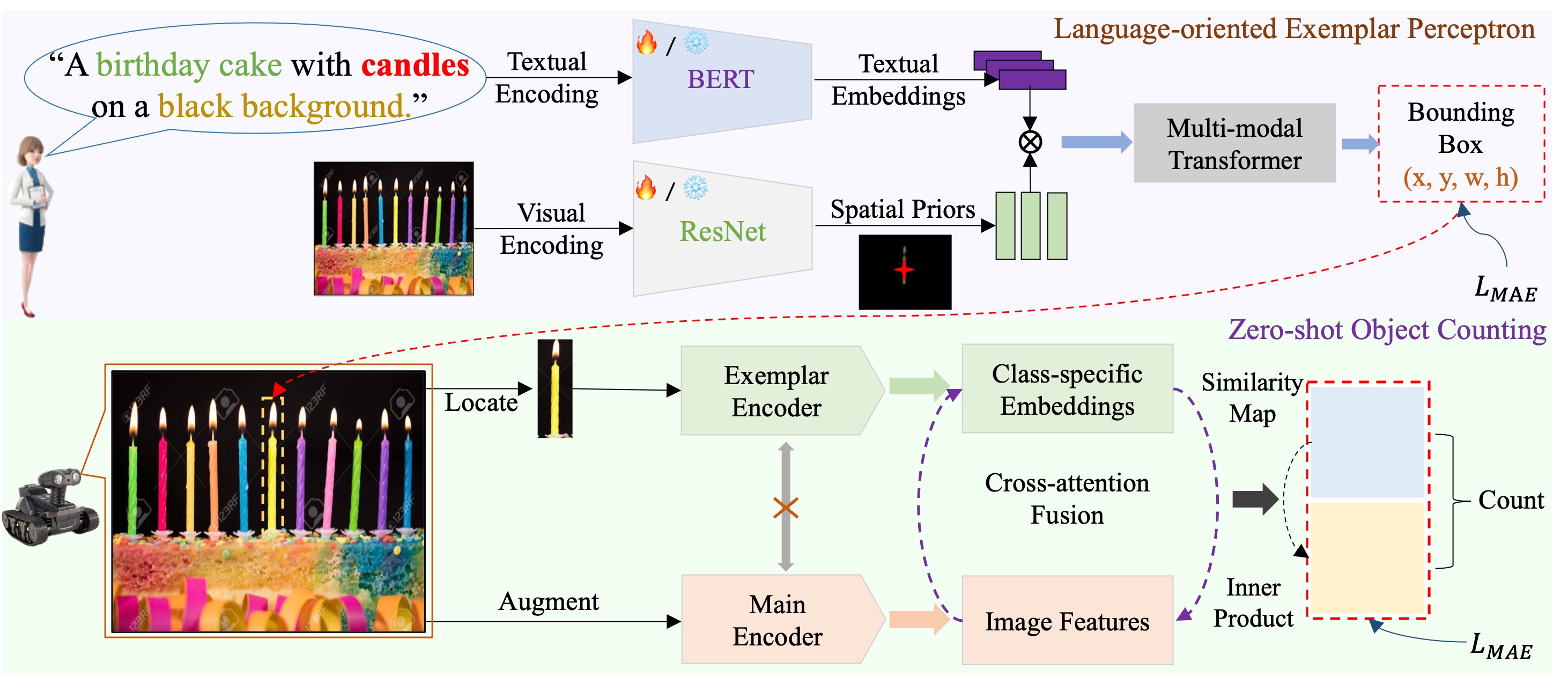}
	\end{center} 
	\caption{The overall architecture of our ExpressCount, which introduces an effective language-oriented exemplar perceptron into visual counting tasks. Specifically, the exemplar perceptron takes a natural image and one detailed expression as input, characterizing and blending both textual and visual signals to guide the accurate learning of exemplars. Moreover, a dual-branch network with cross-attention recalibration is proposed to automatically perform similarity learning, finally inferring the counts of object instances with unseen classes.}
	\label{fig:main}
\end{figure*}

\begin{itemize}
	\item  \textbf{In Use:} A novel \emph{ExpressCount} is presented to guide the learning of accurate exemplar cues (\eg object locations and scales) by delving deep into language-oriented semantic priors. Furthermore, we delicately devise a dual-branch network to execute fine-grained similarity learning, thereby enhancing zero-shot counting performance.
	\item \textbf{Application Scope:} The proposed perceptron embraces the merits of great flexibility and broad applicability. For example, it can be easily deployed in other exemplar-unknown counting scenarios without the need for further troublesome fine-tuning. Our language-centered attempts contribute to the advancements of human-machine interaction and intellectualization of visual counting models.
	\item \textbf{SOTA Performance:} The new ExpressCount surpasses all CAC approaches, yielding state-of-the-art counting performance, and even demonstrates accuracies on par with approaches depending on user-provided exemplars. 
	\item \textbf{Evaluation Venue:} The prevailing and large-scale CAC-specific FSC-147 dataset is specifically extended to blaze a new trail (FSC-147-Express) for developing and assessing the promising language-guided counting models.
\end{itemize}

%% file: 2_review.tex
\section{Related Work}
Recently, numerous Class-Specific Counting (CSC) approaches~\cite{liang2023crowdclip,ranasinghe2023diffuse,song2021rethinking,li2018csrnet,khan2016deep,rahnemoonfar2017deep,bui2020vehicle} have been raised to count object instances with pre-specified categories. While these approaches demonstrate impressive performance, their success hinges on the troublesome task of amassing sufficient samples for each object category. This hurdle impedes the practical applicability of  counting models in real-world scenarios, leading to the surge in popularity of Class-Agnostic Counting (CAC). The realm of CAC can be roughly subdivided into three sub-fields: Exemplar-dependent Counting (EDC), Implicit Exemplar Learning (IEL) and Generalized Exemplar Learning (GEL).
\vspace{-0.5cm}
\paragraph{Exemplar-dependent Counting (EDC).} In an effort to mitigate the burdens associated with acquiring massive training images for all object classes, EDC aims to generate high-fidelity similarity maps by utilizing user-provided exemplars and raw images in a few-shot manner. The pioneering GMN~\cite{lu2018class} reformulates the counting task as the behaviour of learning similarity between exemplar features and sub-patches across  image feature maps.  CFOCNet~\cite{yang2021class} and FAMNet~\cite{ranjan2021learning} enhance the efficacy of few-shot counting models by incorporating more sophisticated feature extraction modules. It is worth noting that FAMNet introduces the first widely-adopted CAC-specific dataset, FSC-147. This popular dataset has sparked a new wave of contemporary CAC, thanks to the availability of extensive natural samples covering a wide range of object classes, such as CounTR~\cite{liu2022countr}, LOCA~\cite{djukic2022low}, SPDCN~\cite{lin2022scale}, BMNet~\cite{shi2022represent} and SAFECount~\cite{you2023few}. Albeit enabling the counting of arbitrary categories, these methods inevitably impose a new requirement for at least one user-provided exemplar (few-shot), which is usually beyond attainment in many real-world applications, such as monitoring system in the wild. 
\vspace{-0.5cm}
\paragraph{Implicit Exemplar Learning (IEL).}
To surmount the drawbacks of EDC, fully zero-shot CAC is catching increasing attention due to its emphasis on heightened generalization \& intellectualization abilities in the field of visual counting. Specifically, RepRPN-Counter~\cite{ranjan2022exemplar} takes the first step to achieve zero-shot optimization by capturing the salient object from the original scenes via a region proposal module. Simultaneously,  RCC~\cite{hobley2022learning} and GCNet~\cite{wang2023gcnet} strive to directly identify exemplars from raw input images by devising multifarious learning units to implicitly extract repetitive patterns. More recently,  CLIP-Count~\cite{jiang2023clip}  integrates the widely-used CLIP model~\cite{radford2021learning} into the counting problems with the goal of utilizing class name (\eg apple and pen) to enhance zero-shot counting capabilities. Despite the diverse strategies designed by previous endeavours to automate exemplar learning, they still deliver unsatisfactory performance as the existing guidance signals, such as raw image and class name, fall short in providing richer semantic information for exemplar localization.

\vspace{-0.5cm}
\paragraph{Generalized Exemplar Learning (GEL).}
Although IEL undeniably contributes to the advancement of zero-shot counting models, the exemplar mining is entangled with the pivotal counting procedure. This severely degrades the accuracy and flexibility of exemplar learning, primarily attributed to the propagation of noise from the counting optimization branch. Therefore, several counting approaches present the strategies for generalized exemplar learning, leveraging the advantageous play-and-plug attribute. For example, 
the latest ZSC~\cite{xu2023zero} utilizes a name-conditional Variational AutoEncoder (VAE) model on close-set classes to explicitly predict the exemplar patches. Despite attaining impressive results, using time-consuming transversal algorithm for selecting optimized patches poses a severe impediment to the model's deployment efficiency. Analogous to CLIP-Count, ZSC also adopts one-word name as the signal to guide the searching of target patches, ignoring the positive influences of rich semantic priors embedded in linguistic expressions. In the coarse of building our FSC-147-Express, a concurrent attempt~\cite{amini2023open} aimed at augmenting the original FSC-147 has been made by simply adding words on class names for each sample (varied images may correspond to the same description). However, we empirically observe that utilizing its dataset~\cite{amini2023open} yields poorer accuracy compared to using off-the-peg
class labels. On the contrary, we meticulously annotate each image with fine-grained descriptions, imbuing them with rich semantics.

%In comparison, our ExpressCount possesses advantages of both higher-efficiency (\ie discarding traversal overheads) and strong interaction. Instead of relying on one-word class name in CLIP-Count and ZSC, we extensively explore the human-imitated expression comprehension imposed on counting problems to pave the way for adopting modern language-vision models into the counting tasks. Towards better validating impacts of natural language on counting models, we also release an extended version of FSC-147 with fine-grained referring expressions. Concurrent with our attempts,  FSC-147-D~\cite{amini2023open} presents textual descriptions with correction and adjustment on class names. However, we observe that adopting descriptions from FSC-147-D degenerates the performance than using the original class names in FSC-147. In contrast, we delicately annotate referring expression with fine-grained semantics for each image in FSC-147.

%% file: 3_method.tex
\section{Methodology}
As depicted in Figure~\ref{fig:main}, the overall architecture of our ExpressCount is mainly comprised of \emph{a novel Language-oriented Exemplar Perceptron}, succeeded by a \emph{Zero-shot Object Counting} branch to perform zero-shot counting. %Specifically, when presented a textual expression coupled with the target scene, our perceptron excels at comprehending the natural language with rich semantic priors. On top of the derived exemplar hints, the subsequent dual-branch counting network proceeds to infer the total number of instances with unseen categories. The pipeline of ``language parsing$\rightarrow$zero-shot counting" marries the counting framework with human-imitated ability.
\vspace{-0.2cm}
\subsection{Language-oriented Exemplar Perceptron}
The past few years have witnessed the flourishing of Large Language Models (LLMs) in many visual problems, such as point cloud analysis~\cite{zhu2023pointclip,wu2023sketch,liu2022towards,10056404,han20223crossnet} and  visual grounding~\cite{zhu2022seqtr,su2021stvgbert,guo2023viewrefer}. Thanks to strong semantic priors, LLMs substantially advance the capabilities, such as intelligence and interaction, across various vision problems. Considering the backwardness of language-based priors in the field of object counting, we pioneer to learn sufficient exemplar hints (\eg location and scale) through the high-quality guidance of natural language. In particular, a language-oriented exemplar perceptron is proposed to characterize exemplar information, including geometries, scales and locations, by learning a mapping between textual expression and bounding box of target exemplar. Concretely, the perceptron consists of a \emph{Linguistic Encoder}, a \emph{Visual Encoder} and a \emph{Transformer-based Multi-modal Integration} module, which will be elaborated in the rest of this section. 

\vspace{-0.5cm}
\paragraph{Linguistic Encoding.}Taking a language expression/text prompt $re_l$ as input, we first utilize the basic pre-trained BERT model~\cite{devlin2018bert} as the LLM to extract rich semantics from linguistic signals, since modern LLMs are often evolved from the fundamental BERT. In specific, the extraction of linguistic representation is achieved by the sequence of tokenization ($T_l$) and embedding ($E_l$) operations. Wherein, $T_l$ is responsible for converting every word from the given sentence into diverse tokenized features (tokens) in the vocab, whereas $E_l$ units are built upon a set of transformer layers, aiming to capture high-level semantics from tokens. Following the common practice of BERT-based models, a randomly-generated token \emph{[Loc]} is introduced as the regression token and is concatenated with the tokenized features, while a token \emph{[sep]} is placed at the end of the tokenized features. The amalgamated  token sequence is fed into the embedding units to model the global relationships among tokens. Since the LLM was pre-trained by hundreds of millions of natural language instances, its generated embeddings effortlessly distill and encapsulate rich semantic priors from the input expressions. The calculation can be formulated as $F_l = E_l(T_l(re_l)) \in R^{D_l\times N_l}$, where $D_l$ and $N_l$ denote the length and number of the language tokens, and are set to 768 and 20, respectively. For simplicity, the dimension of batch size is omitted in presenting shapes.

\vspace{-0.5cm}
\paragraph{Visual Encoding.} Although the pre-trained LLM proves advantageous in mining adequate structural information from linguistic expressions, such as approximate  object locations and instance distributions, its deficiency in spacial guidance undermines the learning quality of exemplars. Therefore, a parallel visual encoding branch is introduced, tasked with processing a target image as its input, $x_i \in R^{3 \times H_l \times W_l}$. Specifically, the visual encoding workflow is comprised of shallow feature extraction and high-level semantics modeling.  A ResNet-like encoder ($R_i$) is first employed to extract shallow representations with sufficient detailed information (such as texture and contour), denoted as $f_{s} \in R^{2048\times \frac{H_l}{32} \times \frac{W_l}{32} }$, which is subsequently fed into a 1$\times$1 convolution layer to compress the 2048 kernels into compact 256-channel features $f_{st} $. A set of cascaded transformer layers (the width of 256) $T_i$ with 8 self-attention heads are finally incorporated to model high-level semantics with global receptive fields. The visual encoding procedure is denoted as $F_i = T_i(R_i(x_i)) \in R^{D_i\times N_i}$, where $D_i=256$ represents channel length of output embeddings whereas $N_i=\frac{H_l}{32}\times \frac{W_l}{32}$ denotes the token number.

%and force the holistic model to regress the target coordinates conditioned on the input expression $C(F_l|x_i)$. The  visual encoding workflow can be divided into shallow feature extraction and global semantics modeling.
\vspace{-0.5cm}
\paragraph{Multi-modal Integration and Regression.}

%\begin{figure}[h]
%	\begin{center}
	%		\includegraphics[width=\linewidth]{LV.pdf}
	%	\end{center} 
%	\caption{The details of our proposed multi-modal integration pattern for location prediction from linguistic inputs.}
%	\label{fig:LV}
%\end{figure}
At the end of language-oriented exemplar perceptron, a transformer-based integration module is designed to blend the hybrid modalities of textual and visual embeddings. Thanks to the higher capability for characterizing global semantics, the multi-modal transformer used in our method is competent to regress the accurate exemplar bounding boxes. Specifically, a concatenation is performed on a learnable token $[Loc] \in R^{256\times 1}$ and text-vision embeddings [$Loc$, $F_l$, $F_i$] $\rightarrow$$F_{li} \in R^{256\times (1+N_l+N_i)}$. Then $F_{li}$ is fed into a sequence of 6 transformer layers to mix and characterize the multi-modal representations, thereby imbuing the regression token $[Loc]$ with rich semantics derived from both linguistic and visual signals. The final exemplar regression head involves ``three Linear Projections, one Sigmoid Function", which takes as input the learned $[Loc]$ and generate the bounding boxes of exemplar with the shape of $(x,y,h,w)$. %To better understand the aforementioned procedure, we visualize the integration scheme in Figure~\ref{fig:LV}.
\vspace{-0.2cm}
\subsection{Zero-shot Object Counting}
%Human-imitated expression comprehension attempts to understand spatial distributions of objects and infer the possible location cues by assimilating easy-to-get natural language signals, which lays the foundation of the zero-shot counting in this paper. 

%\vspace{-0.2cm}
\paragraph{Dual-branch Backbone.}
Different from single encoder for both exemplar and raw images in previous methods~\cite{lin2022scale,shi2022represent,xu2023zero}, we introduce a dual-branch backbone for the respective extraction of features from the input exemplar and raw image, as we empirically observe a diversity in receptive fields and semantics (object details \& instance repetition) between images that encompass single and multiple object instances. Beginning with the raw image $I\in R^{3\times H \times W}$ and the learned bounding box $(x,y,h,w)$, the exemplar patch $I_e \in R^{3\times128\times128}$ is obtained by cropping and resizing from the original $I$ automatically. Subsequently, the exemplar, along with the raw image, is fed into the dual-branch ResNets ($R_e$ and $R_I$) to calculate basic 2D features $F_e=R_e(I_e) \in R^{2048\times8\times8}$ and $F_I=R_I(I_I) \in R^{2048\times\frac{H}{16}\times \frac{W}{16}}$. Several 1$\times$1 convolution layers are then imposed to reduce and align the channel dimension, and reshape the representations as $F_e\in R^{256\times8\times8}$ \& $F_I \in R^{256\times\frac{H}{16}\times \frac{W}{16}}$. It is worth noting that, to represent 1D global semantics from the exemplar for effective similarity learning, we delicately design a fine-grained extractor comprised of a sequence of ``Patch Splitting$\rightarrow$Linear Projection$\rightarrow$1$\times$1 Conv", instead of Average Global Pooling (AVG) adopted by existing approaches~\cite{shi2022represent,xu2023zero}. By doing so, a vector $\hat{F_e}\in R^{1\times256}$ with higher-level semantic information of exemplar patches is computed and refined. 

\vspace{-0.5cm}
\paragraph{Cross-attention Recalibration.}
Upon the transformation of the exemplar and raw image into preliminary features ($\hat{F_e}$ and $F_I$), we further advance to execute the learning of high-fidelity similarity map through a cross-attention mechanism. Due to the synergy between exemplar and dense instances in feature space, a cross-attention scheme $CA(\cdot)$ is proposed to recalibrate the features conditioned on each other, denoted as $CA(\hat{F_e}|F_I) \cup CA(F_I|\hat{F_e})$. Specifically, to enrich the exemplar feature $\hat{F_e}\in R^{1\times256}$, the $CA$ function takes image feature $F_I \in R^{256\times\frac{H}{16}\times \frac{W}{16}}$ as the input and utilize a set of operations ( ``Matrix Multiplication $M_1$, Linear Transformation $L_1$ and Sigmoid Function") to deduce the recalibration weights $W_1\in R^{1\times256} =L_1(M_1(\alpha_1F_I,\alpha_2F_I)).Sigmoid()$, where $\alpha_1$ and $\alpha_2$ are the trainable balance factors. In the opposite direction, the input $\hat{F_e} \in R^{1\times256}$ passes through the calculation units of ``Linear Transformation $L_2$ and Sigmoid Function" towards generating the weights $W_2\in R^{1\times256}=L_2(\hat{F_e}).Sigmoid()$. The extracted weights $W_1$ and $W_2$ are employed for channel-wise recalibration of coarse features, expressed as  $\hat{F_e}  = \hat{F_e}  \times W_1$ and $F_I = F_I \times W_2^T$.

\vspace{-0.5cm}
\paragraph{Counting-specific Expression Annotating.}
\begin{figure}[h]
	\begin{center}
		\includegraphics[width=\linewidth]{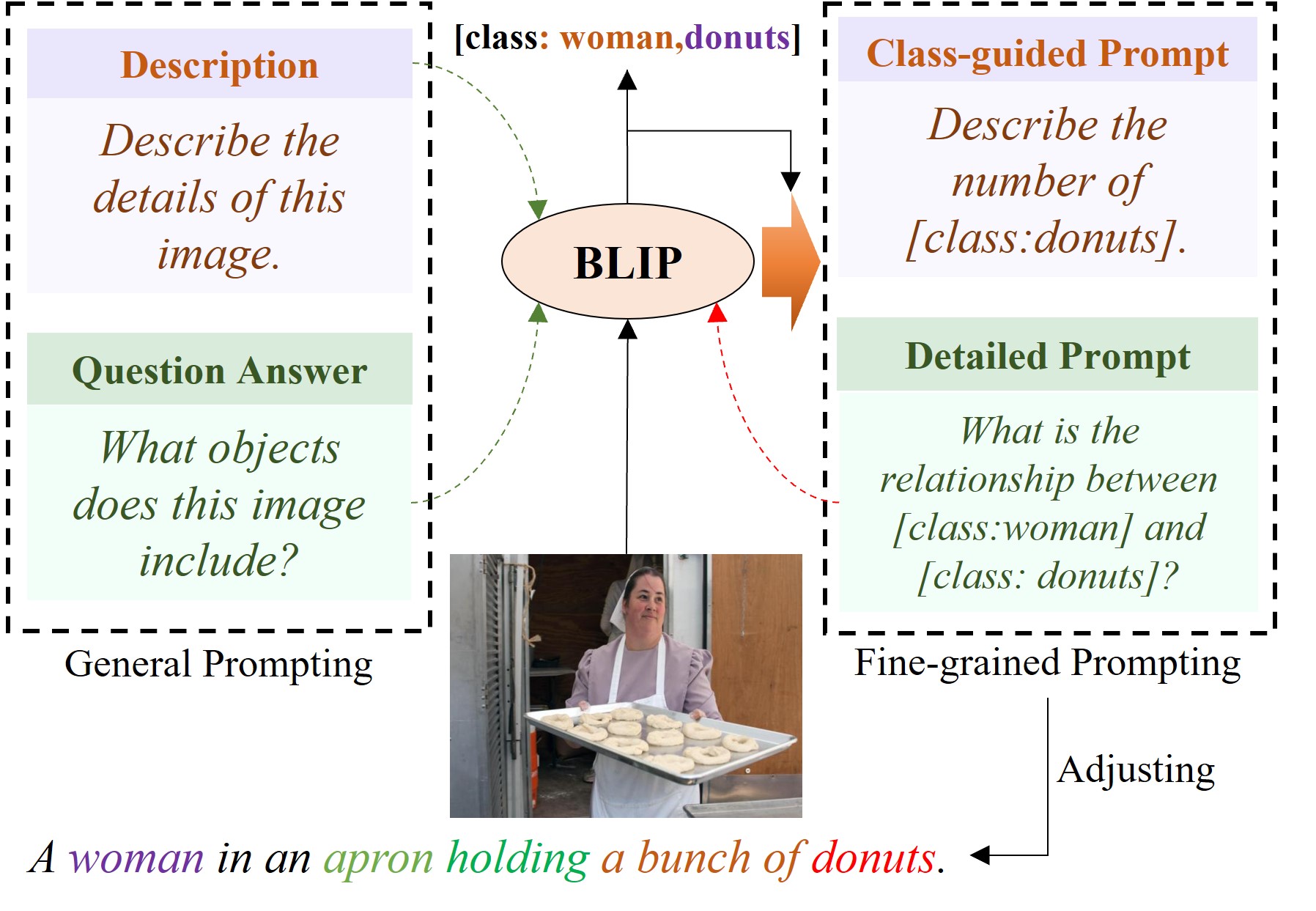}
	\end{center} 
	\caption{The generation example of annotating language expressions in the language-vision counting dataset (FSC-147-Express).}
	\label{fig:expressgen}
\end{figure}
Despite the abundance of language-image data for guiding many vision tasks, such as image synthesis, investigating into the impacts of language on counting problems remains scarce, primarily attributed to the absence of language-vision annotations. This dearth hampers the interaction and generalization capabilities of current CAC models. To address this limitation, in this paper, we propose a counting-specific language-vision dataset for training and evaluating language-based counting methods, namely \emph{FSC-147-Express}, by extending the most prevailing FSC-147. As depicted in Figure~\ref{fig:expressgen}, we aspire to distill valuable knowledge relevant to the given counting scene from the pre-trained BLIP. For example, when presented a counting scene, high-level prompts are designed (\eg \emph{``What objects does this image include?"}) to generate coarse answers involving key object class names. Based on these coarse hints, we further refine them by leveraging various counting-specific prompts (\eg \emph{``How many donuts? Describe the [class] details."}). Consequently, a series of fine-grained expressions are outputted such as quantifier (``a bunch of/a bowl of") and action (``holding"). Through the above meticulous adjustments, implicit semantic priors  (\ie \emph{``a bunch of"} and \emph{``a pile of"}) are inherited from the BLIP model to enhance the discrimination of images with varying levels of instance density.

\vspace{-0.5cm}
\paragraph{Optimization Objective.}
The entire optimization of our ExpressCount depends on hybrid supervisory signals from both language-oriented exemplar perceptron and zero-shot counting branch. First, in adherence to prevalent practices in object detection, the $L_1$ norm is utilized to calculate the supervisory signal $L_l$, quantifying the distance between predicted bounding boxes $E_p(x_p,y_p,h_p,w_p)$ and their corresponding ground truths $E_g(x_g,y_g,h_g,w_g)$ (taking the prediction of a single exemplar as an example):
\begin{equation}\label{LE}
	L_l = \sum_{i=1}^{N}(|x_g-x_p)|+|y_g-y_p|+|h_g-h_p|+|w_g-w_p|),
\end{equation}
where $N$ denotes the batch size whereas $h_i, i\in[p,g]$  and $w_i,i\in[p,g]$ indicate the heights and widths of bounding boxes, respectively. Simultaneously, the second loss term $L_c$ is imposed on the dual-branch counting sub-network to measure the counting error between predicted and ground-truth counts $C_p$ and $C_g$:
\begin{equation}\label{LE}
L_c = \sum_{i=1}^{N}(|C_g-C_p|),
\end{equation}
where $N$ is the batch size and the total objective function is formulated as $L=L_l+L_c$.

%% file: 4_experiments.tex
\section{Experiments}

\begin{table*}[h]
	\begin{center}
		\setlength{\tabcolsep}{3.6mm}
		\begin{tabular}{ccccccc}
			\hline
			Frameworks & Exemplar & Type & {\bf MAE} (\emph{Val})  & {\bf RMSE} (\emph{Val}) & {\bf MAE} (\emph{Test}) & {\bf RMSE} (\emph{Test})\\
			\hline
			GMN~\cite{lu2018class} & \checkmark & EDC & 29.66 & 89.81 & 26.52 & 124.57 \\
			FamNet~\cite{ranjan2021learning} &  \checkmark & EDC & 24.32 &70.94 & 22.56 & 101.54 \\
			FamNet+~\cite{ranjan2021learning} &  \checkmark & EDC & 23.75 & 69.07 & 22.08 & 99.54\\
			CFOCNet~\cite{yang2021class} &  \checkmark & EDC & 21.19 & 61.41 & 22.10 & 112.71\\
			BMNet~\cite{shi2022represent} &  \checkmark & EDC & 19.06 & 67.95 &  16.71 & 103.31\\
			CountTR~\cite{liu2022countr} &  \checkmark & EDC & {\bf 13.13} & {\bf 49.83} & {\bf 11.95} & {\bf 91.23}\\
			\hline
			RepRPN-Counter~\cite{ranjan2022exemplar}  &  $\times$ &  IEL & 30.40
			& 98.73 & 27.45 & 129.69 \\
			GCNet~\cite{wang2023gcnet}  &  $\times$ &  IEL & 19.50 & 63.13 & 17.83 & {\bf102.89} \\
			CLIPCount~\cite{jiang2023clip}  &  $\times$ &  IEL & {\bf 18.79} & {\bf61.18} & {\bf17.78} & 106.62 \\
			\hline
			ZSC-RPN~\cite{xu2023zero}  &  $\times$ &  GEL & 32.19 & 99.21 & 29.25 & 130.65 \\
			ZSC-PatchSelection~\cite{xu2023zero}  &  $\times$ &  GEL & \fbox{26.93} & \fbox{88.63} & \fbox{22.09} & \fbox{115.17}\\
			ExpressCount-E(ours) &  $\times$ &  GEL & 19.84 & 60.40 & 21.32 & 110.30 \\
			%ExpressCount-V(ours) &  $\times$ &  GEC & 19.18 & 64.92 & 19.08 & 108.29 \\
			ExpressCount-EV(ours) & $\times$ & GEL & 19.04 & 56.72 & 20.42 & 107.99 \\
			ExpressCount(ours) &  $\times$ &  GEL & {\bf17.33} & {\bf54.00} & {\bf18.92} & {\bf101.20}\\
			\hline
		\end{tabular}
	\end{center}
	\caption{ The quantitative comparisons on the \emph{Val} and \emph{Test} sets of FSC-147 and FSC-147-Express. The best results for three types, Exemplar-Dependent Counting (EDC), Implicit Exemplar Learning (IEL) and Generalized Exemplar Learning (GEL),  are highlighted in boldface, which demonstrates that our method achieves state-of-the-art performance compared with existing exemplar-free methods.}
	\label{table:comparison}
\end{table*}
\vspace{-0.2cm}
\paragraph{Implementation Details.}
\emph{i)} For the Language-oriented Exemplar Perceptron, the input image is resized to 640$\times$640 and the length of expression token is fixed as 20. The linguistic and visual encoders consist of 12 and 6 transformer layers, respectively, while additional 6 layers are stacked sequentially in multi-modal integration and regression module. %We initialize the linguistic part using a pre-trained BERT parameters to enrich semantic priors while surmounting the drawback of limited language labels in FSC-147-Express. 
To expedite the convergence of the exemplar perceptron, the visual encoder is initialized with parameters from the counterpart in TransVG~\cite{deng2021transvg}; \emph{ii)} For the zero-shot counting branch, the augmentation of random reshaping \& erasing is employed to mitigate the risk of potential overfitting.  During the training phase, the learning rate and batch size are chosen as 1e-5 and 15 respectively. The ExpressCount is optimized in an end-to-end manner using the AdamW optimizer with a weight decay of 5e-4.

\vspace{-0.5cm}
\paragraph{Dataset and Evaluation Metrics.} The FSC-147-Express and all comparisons are based on the largest-scale FSC-147 benchmark, widely adopted by existing CAC approaches. This dataset comprises total 6,135 images sourced from the Internet, including 147 different object categories spanning from candies and comic books to apples. It is subdivided into three subsets for training (3,659 samples), validating (1,286 samples), and testing (1,190 images) models. Mean Absolute Error (MAE) and Root Mean Square Error (RMSE) are incorporated as evaluation metrics:  $MAE=\frac{1}{B}\sum_{i=1}^{B}|C_p-C_g|$ and $RMSE=\sqrt{\frac{1}{B}\sum_{i=1}^{B}(C_p-{C_g}^2)}$, where $B$ is the number of testing samples, whereas $C_p$ and $C_g$ indicates the predicted and ground-truth instance counts. 
In this paper, three versions of our language-oriented counting model are developed and compared, including \emph{(1)} {\bf ExpressCount-E}, where the exemplar perceptron undergoes training exclusively through language expressions, devoid of any reliance on image signals; \emph{(2)} {\bf ExpressCount-EV}, a refinement of ExpressCount-E, this variant incorporates annotated language-image pairs, but with a frozen visual encoder to enrich spatial semantics; and \emph{(3)} {\bf ExpressCount}, which further delves deep into the language-oriented semantic priors through the collaborative fine-tuning of linguistic and visual encoders.
\vspace{-0.2cm}
\subsection{Comparisons with State of the Arts}

\begin{figure}[!h]
	\begin{center}
		\includegraphics[width=\linewidth]{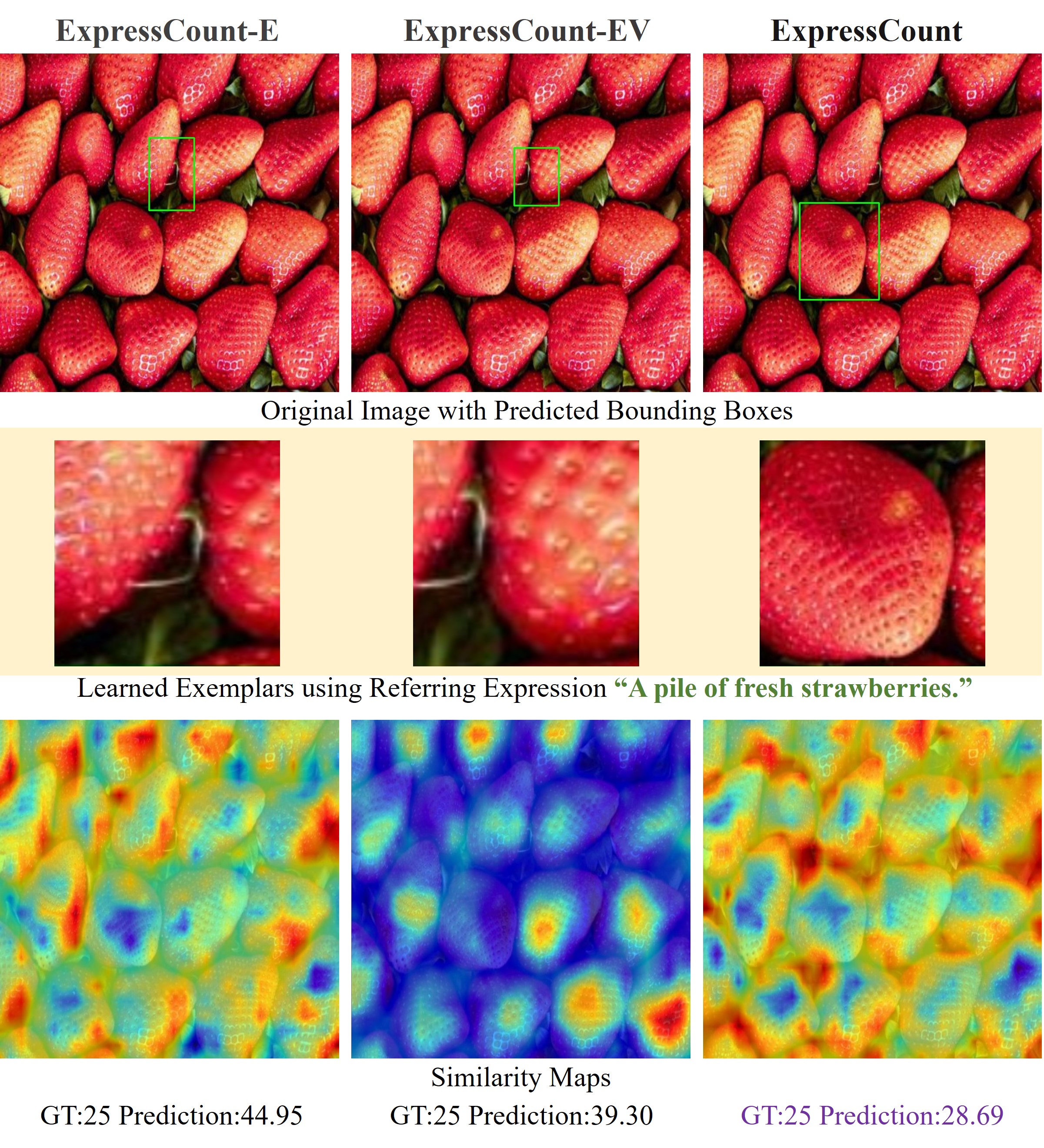}
	\end{center} 
	\caption{Visual comparison among three versions of our language-based methods. The first row depicts the exemplar predictions, the second row shows the captured exemplar patches in zoomed-in views, and the third row demonstrates three corresponding similarity maps. It can be observed that the guidance of language- and image-oriented semantic priors both contributes to the enhance the accuracy of exemplar prediction.} 
	\label{fig:3V}
\end{figure}
Table~\ref{table:comparison} provides a comprehensive comparison between ExpressCount variants and existing CAC approaches, which demonstrates that the zero-shot counting performance is indeed enhanced by endowing the model with rich language-oriented semantic priors. In particular, our nascent ExpressCount-E outperforms the SOTA ZSC model~\cite{xu2023zero} in the scope of GEL approaches by a large margin. It attains a considerable decrease of 26.32\% in MAE and 31.85\% in RMSE for the validation set, along with a decrease of 3.48\% in MAE and 4.22\% in RMSE for the test set. Furthermore, the ExpressCount-EV extends upon this basic model, enhancing the efficacy of language-guided exemplar learning.  The accuracies stand at 19.04 MAE and 56.72 RMSE for the validation set, and 20.42 MAE and 107.99 RMSE for the test set, surpassing a series of popular exemplar-dependent approaches, including GMN, FamNet, and CFOCNet. Finally, the ExpressCount model excels among existing EFC algorithms by co-optimizing the linguistic and visual encoders presented in the stem of language-oriented exemplar perceptron, resulting in state-of-the-art  MAE of 17.33 and RMSE of 54.00 for the validation set, as well as the RMSE of 101.20 for the test set. Particularly, within the context of GEL, it attains a substantial improvement of 35.64\% and 39.07\% lower MAE and RMSE for the validation set, whereas 14.35\% and 12.12\% lower MAE and RMSE for the test set. When compared with inefficient and low-applicability EDC models, our ExpressCount even secures the second position in terms of accuracy on the validation set, trailing only behind CountTR (MAE of 13.13 and RMSE of 49.83). To visually illustrate the difference among results produced by our three versions, we offer an example from the test dataset in Figure~\ref{fig:3V}.

\vspace{-0.2cm}
\subsection{Ablation Studies}

\begin{figure}[h]
\begin{center}
	\includegraphics[width=\linewidth]{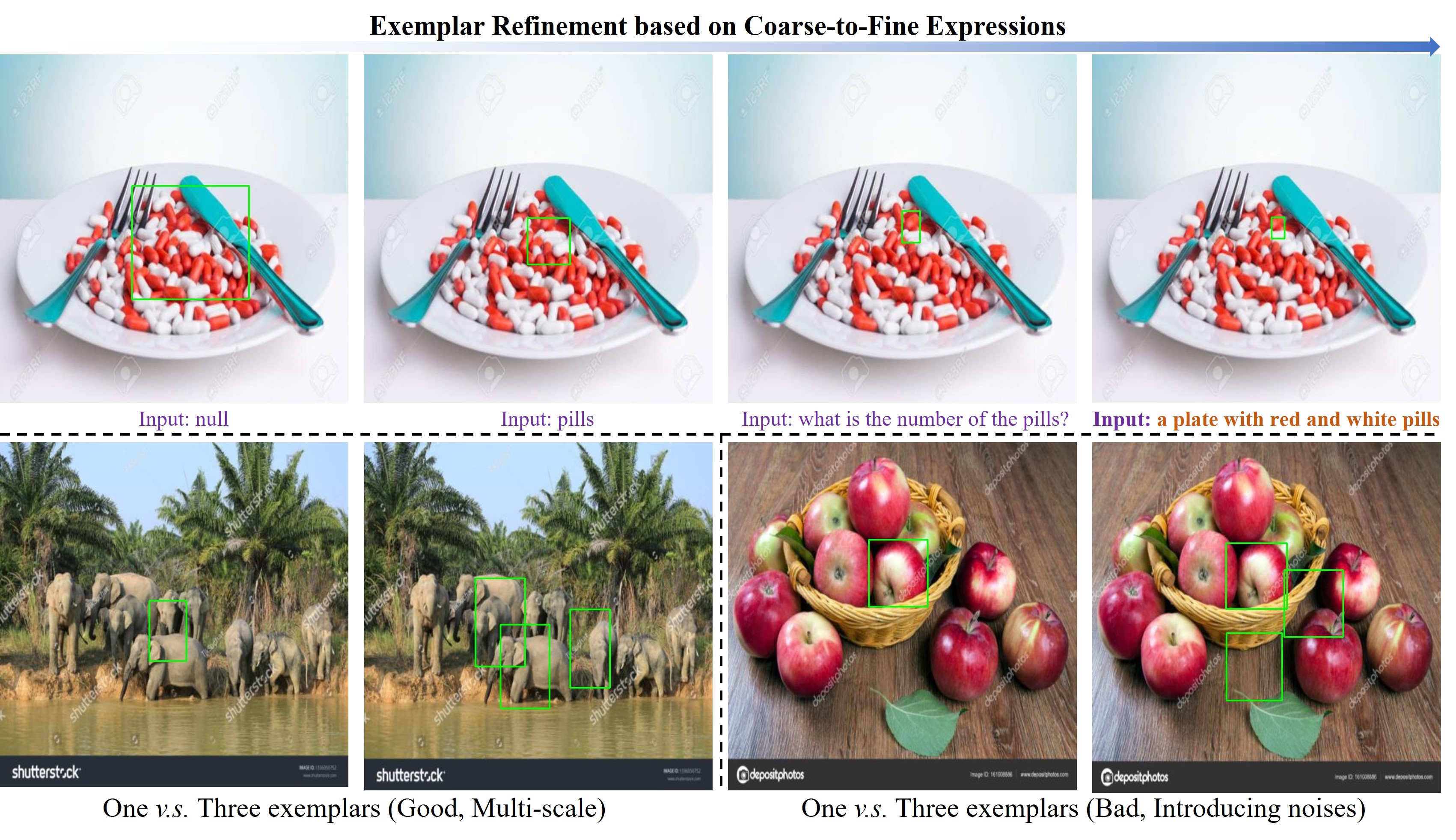}
\end{center} 
\caption{Visual comparisons for ablation studies. Top: exemplar extraction under coarse-to-fine expressions shows the benefit of detailed language expressions. Bottom: predicting three exemplars instead of one may have both positive (diversified samples) or bad (added noises) impacts.} 
\label{fig:MComp}
\end{figure}

\paragraph{The Effects of Varying Expressions.}
We start with analyzing the influences of different prompt types (such as \emph{NULL, class name}) on the zero-shot counting, see Table~\ref{table:abloss}. Among these studies, a visual encoder-frozen ExpressCount-EV is adopted as the baseline, enabling the isolation of effects attributed to fine-tune the visual encoder. Quantitative results illustrate that, as the expressions become more intricate and detailed, the precision of zero-shot counting significantly enhances. This phenomenon can be elucidated by the fact that detailed expressions allow for inheriting richer semantics from the LLM, thereby guiding the effective and accurate exemplar prediction, see Figure~\ref{fig:MComp}.

\begin{table}[h]
	\begin{center}
		\setlength{\tabcolsep}{2.5mm}
		\begin{tabular}{ccc}
			\hline
			\multirow{2}{*}{Language Prompts}& \multicolumn{2}{c}{{\bf \emph{Val Set}}} \\ 
			~ & MAE & RMSE\\
			\hline
			NULL   &  24.16 & 75.44  \\
			``[class name]" & 21.78 &  63.52 \\
			``What is the number of the [class]?" &  20.96 & 60.48  \\
			Our Fine-grained Expressions & {\bf19.04} & {\bf 56.72} \\
			\hline
		\end{tabular}
	\end{center}
	\caption{Ablation studies on impacts of varying expressions types.} %The best performer across model and loss type per metric is bolded.}
\label{table:abloss}
\end{table}

\vspace{-0.5cm}
\paragraph{The Guidance of One \emph{v.s.} Three Exemplars.} 
The majority of existing CAC approaches typically address scale variations by offering multiple exemplars (usually three). Therefore, we explore the language-guided prediction of multiple exemplars. Following the choice in ZSC~\cite{jiang2023clip} and BMNet~\cite{shi2022represent}, we re-implement the regression head in our exemplar perceptron  by adding more learnable token heads, such as $[Loc_1, Loc_2, Loc_3]$. The comparison results are reported in Table~\ref{table:MEL}, while the visual outcomes are provided in Figure~\ref{fig:MComp} (second row). Observing the results, it becomes evident that the inclusion of multiple exemplar predictions yields improvements in the RMSE metric, albeit at the expense of somewhat deteriorated MAE (may be caused by the noisy exemplar predictions).

\begin{table}[h]
\begin{center}
	\setlength{\tabcolsep}{2.1mm}
	\begin{tabular}{ccccc}
		\hline
		\multirow{2}{*}{Exemplar Number}& \multicolumn{2}{c}{{\bf \emph{Val Set}}} & \multicolumn{2}{c}{{\bf \emph{Test Set}}} \\ 
		~ & MAE & RMSE & MAE & RMSE\\
		\hline
		1  &  {\bf17.33} & 54.00 & {\bf18.92} & 101.20\\
		%2  & 18.55 & 57.06 & 20.01 & \\
		3 & 17.40 & {\bf53.31} & 19.16 & {\bf99.60}  \\
		\hline
	\end{tabular}
\end{center}
\caption{Ablation studies on the influences of learning multiple exemplars (one \emph{v.s.} three) on \emph{Val} and \emph{Test} datasets.}
\label{table:MEL}
\end{table}

\begin{figure}[h]
\begin{center}
	\includegraphics[width=\linewidth]{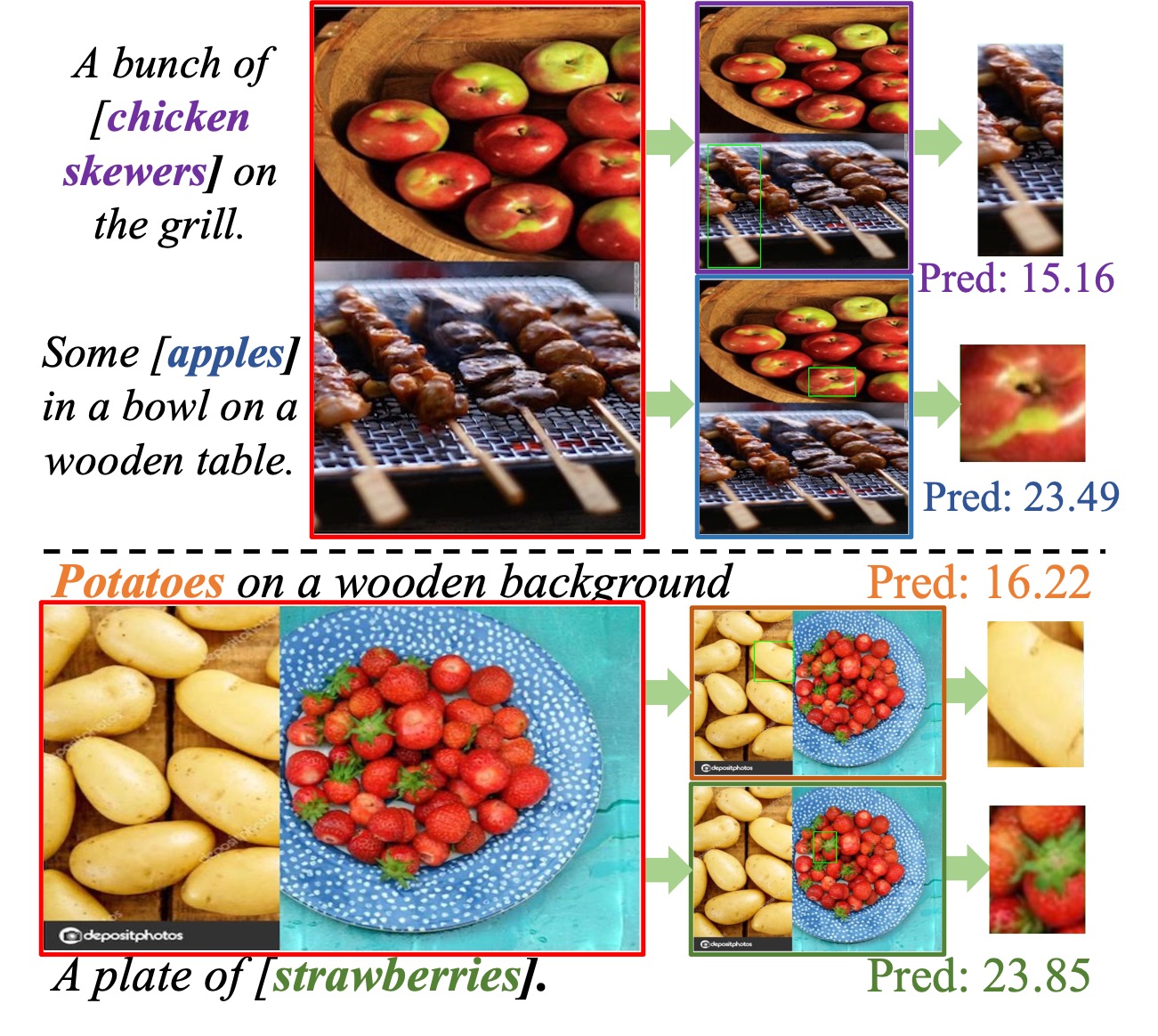}
\end{center} 
\caption{The performance of our model when encountering the scenes with distinct categories.Providing different textual expressions leads to different bounding box prediction results.}
\label{fig:MC}
\end{figure}
%\paragraph{Density Map \vs Count-level Supervisory Signals}

%\begin{table}[h]
%	\begin{center}
%		\setlength{\tabcolsep}{2.0mm}
%		\begin{tabular}{ccccc}
	%			\hline
	%			\multirow{2}{*}{Supervisory Signals}& \multicolumn{2}{c}{{\bf \emph{Val Set}}} & \multicolumn{2}{c}{{\bf \emph{Test Set}}} \\ 
	%			~ & MAE & RMSE & MAE & RMSE\\
	%			\hline
	%			Density Map  &  {\bf17.33} & 54.00 & {\bf18.92} & 101.20\\
	%			Single Counts  &  &  &  & \\
	%			Hybrid  &  & &  & \\
	%			\hline
	%		\end{tabular}
%	\end{center}
%	\caption{Comparisons to explore the influences of training our ExpressCount using different types of supervisory signals. }
%	\label{table:DSC}
%\end{table}

\begin{figure}[!h]
	\begin{center}
		\includegraphics[width=\linewidth]{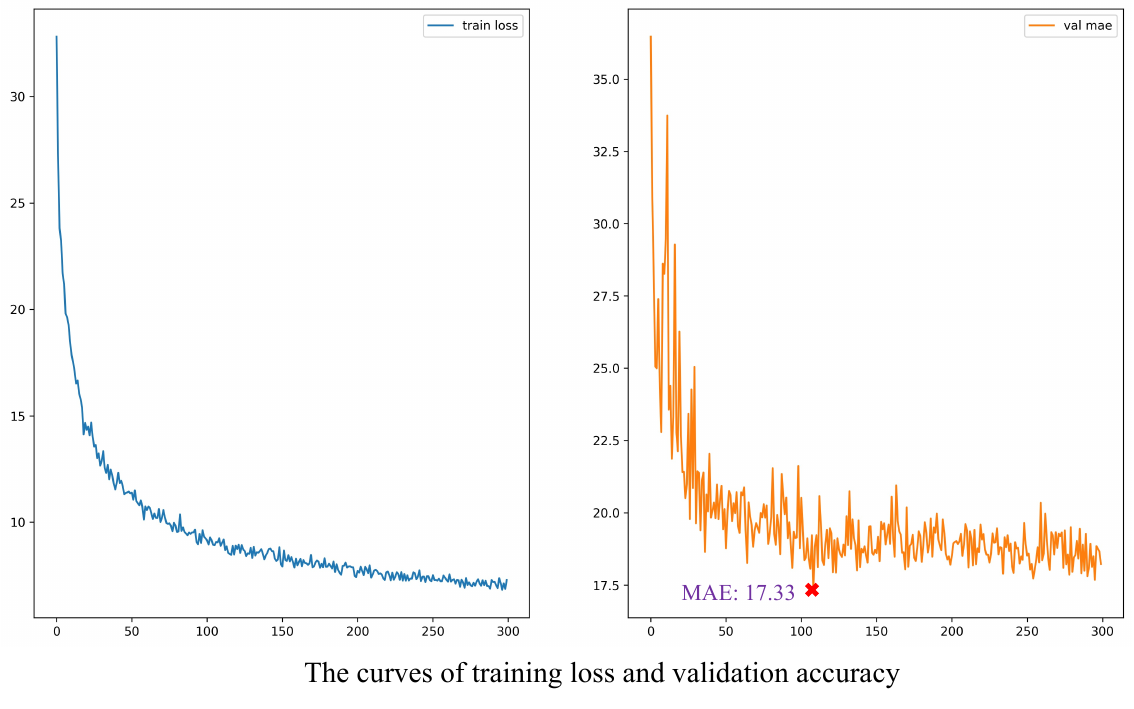}
	\end{center} 
	\caption{The curves depicting loss and val MAE during training.}
	\label{fig:trainingcurve}
\end{figure}

\vspace{-0.5cm}
\paragraph{Exploring the Potential of Addressing Multi-class Scenes.}
Adhering to the studies carried out by ZSC, we further exploit the capability of our ExpressCount to address multi-class scenarios. Specifically, we generate several composite images featuring various object categories, as the availability of natural multi-class samples in the dataset is limited. As depicted in Figure~\ref{fig:MC}, when presented with an image containing two expressions related to different categories, our model allows to predict corresponding exemplar patches, thereby enhancing the inference of infer plausible class-oriented counting results.
\vspace{-0.5cm}
\paragraph{Depicting Curves of Training Loss and Validation Accuracy.}
To visualize the training process, the curves of loss and validation accuracy values are depicted in Figure~\ref{fig:trainingcurve}, which demonstrates the stability of the proposed model.

%\vspace{-0.5cm}
%\paragraph{Visualization Examples on \emph{Test} Set}
%Several prediction examples from \emph{Test} set are visually shown in Figure~\ref{fig:visexamples}. It can be observed that our ExpressCount is capable of learning good exemplars and predicting accurate counts covering a wide range of dense scenes.
%\begin{figure*}[h]
%	\begin{center}
%		\includegraphics[width=0.80\linewidth]{supplyve.pdf}
%	\end{center} 
%	\caption{The visualization examples on \emph{Test} set produced by our ExpressCount involving predicted exemplar boxes and similarity maps, which aims to  assist in demonstrating the robustness of our ExpressNet.}
%	\label{fig:visexamples}
%\end{figure*}

%\subsection{Referring Expressions \emph{v.s.} Class Name}
%To better show the details of the referring expression in our FSC-147-Express, a series of samples are depicted in Figure~\ref{fig:captions}. The schema illustrates that our endeavour allows to produce abundant semantic diversities, such as quantifier (\eg ``a bunch of" and ``a class of") and peripheral object cues (\eg ``different colours" and ``with a racket"), which are beneficial for guiding the inference of the object location cues and the regression of final instance counts.
%\begin{figure}[!h]
%	\begin{center}
%		\includegraphics[width=\linewidth]{caption.pdf}
%	\end{center} 
%	\caption{The visualization examples of expression in our FSC-147-Express compared with the simple class name, which shows that our captions are able to expose more semantic information, such as quantifier semantics.}
%	\label{fig:captions}
%\end{figure}
\vspace{-0.5cm}
\paragraph{Strong Density Maps \emph{v.s.} Weak  Counts.}
In contrast to the recent ZSC model, which utilizes laborious and hard-to-get density maps as supervision, we unprecedentedly introduce the weakly-supervised learning (information-poor single counts) into CAC problems. Through comparisons on \emph{Val} set using density maps and single counts in Table~\ref{table:DSC}, we notably observe that the utilization of strong density maps as supervisory signals adversely affects the counting performance of ExpressCount. The implementation of density map generation follows the practices in existing BMNet and ZSC. This phenomenon illustrates that CNNs-based density prediction head is incapable of capturing global semantic information. In contrast, our single-count regression strategy is imbued with the capacity for global modeling, resulting in heightened robustness to exemplar noises.

%goes against the intuition that the strong pixel-wise constraints can bring huge performance gains. This may be explained by that: \emph{i}) Strong spatial constraints require precise exemplar cues for better matching learning. However, the exemplar prediction procedure inevitably introduces exemplar noise; \emph{ii}) The CNNs-based density prediction structure designed by existing BMNet is incapable of capturing global semantic cues. In contrast, our single-count regression head is imbued with the capacity for global modeling, resulting in heightened robustness to exemplar noises.
\begin{table}[h]
	\begin{center}
		\setlength{\tabcolsep}{2.7mm}
		\begin{tabular}{ccccc}
			\hline
			Supervisory & \multicolumn{2}{c}{{\bf \emph{Val Set}}} & \multicolumn{2}{c}{{\bf \emph{Test Set}}} \\ 
			Signals & MAE & RMSE & MAE & RMSE\\
			\hline
			Single Count  &  {\bf17.33} & 54.00 & {18.92} & 101.20\\
			Density Map  & 21.30 & 79.43 & {18.52} & 125.63\\
			Hybrid & 21.17 & 76.89 & \fbox{\color{blue}{{\bf 16.94}}} & 114.64 \\
			\hline
		\end{tabular}
	\end{center}
	\caption{Ablation studies on the influences of using supervisory signals of Density Map \emph{v.s.} Single Count \emph{v.s.} Their Combination.}
	\label{table:DSC}
\end{table}
In real-world applications, the collection of single-count annotations is easier than density map, so our model possesses wider applicability than all current CAC approaches by evading the notorious process of placing dots on centroids of all object instances. This is a significant merit of counting methods for their development and deployment, especially for class-agnostic counting. Even though the results on \emph{Val} set are degraded, the values of MAE on \emph{Test} set are progressively improved (18.92$\rightarrow$18.52$\rightarrow$16.94). In particular, the MAE value of 16.94 (SOTA) surpasses all existing class-agnostic counting methods by leveraging hybrid supervisory signals (both density maps and single counts). 

\section{Conclusion and Future Work}
A novel framework, \emph{ExpressCount}, is designed in this paper to propel the advancement of zero-shot object counting. One of its distinctive features is the design of language-oriented exemplar perceptron, which contributes to not only yielding significant performance gains but also establishing a crucial connection between modern LLMs and counting tasks. Moreover, an enhanced language-vision dataset (\emph{FSC-147-Express}) is introduced by extending the FSC-147 in vogue to open up a new venue for developing and assessing language-based counting models. In the future,  our focus will shift towards investigating the problem of counting instances with multiple categories through the language guidance. This is a challenge that remains unaddressed by the existing approaches, and we seek to deeply explore this promising avenue by advancing techniques in this paper.